%% file: main.tex
\documentclass[letterpaper, 10 pt, conference]{ieeeconf}
\include{custom_commands.tex}

\IEEEoverridecommandlockouts
\usepackage{amsmath}
\usepackage{amssymb}
\usepackage{float}

\restylefloat{figure}
\title{\LARGE \bf Mobile Robot Localization: a Modular, Odometry-Improving Approach}

\author{Luca Mozzarelli$^{1}$, Luca Cattaneo$^{1}$, Matteo Corno$^{1}$ and Sergio Matteo Savaresi$^{1}$%
\thanks{
$^{1}$ The authors are with the Dipartimento di Elettronica, Informazione e Bioingegneria, Politecnico di Milano, 20133 Milano, Italy.
Email: \texttt{\{luca.mozzarelli, luca1.cattaneo, matteo.corno, sergio.savaresi\}@polimi.it}}%
\thanks{The experimental data presented in this paper was collected during the project \quotes{BUDD-e: Blind-assistive aUtonomous Droid Device}, which is financially supported by Politecnico di Milano, Polisocial Award 2021.}
}

\begin{document}

\maketitle
\thispagestyle{empty}
\pagestyle{empty}

\begin{abstract}
Despite the number of works published in recent years, vehicle localization remains an open, challenging problem.
While map-based localization and SLAM algorithms are getting better and better, they remain a single point of failure in typical localization pipelines.
This paper proposes a modular localization architecture that fuses sensor measurements with the outputs of off-the-shelf localization algorithms.
The fusion filter estimates model uncertainties to improve odometry in case absolute pose measurements are lost entirely.
The architecture is validated experimentally on a real robot navigating autonomously proving a reduction of the position error of more than 90\% with respect to the odometrical estimate without uncertainty estimation in a two-minute navigation period without position measurements.
\end{abstract}

\section{INTRODUCTION}
\label{sec:introduction}
As more and more autonomous mobile robots (AMRs) are getting deployed \quotes{into the wild}, the need for robust, reliable and always-available localization is increasing.
This is especially true when focusing on the urban-environments, where last-mile delivery robots are becoming an interesting option for many businesses, while the localization problem is exacerbated by the challenging context. 
GNSS-based localization with Real Time Kinematics (RTK) corrections has been successfully employed in some application scenarios like agricultural fields, but its quality in urban areas is uneven, with vast areas where the signal has low quality due to multi-path reflections or is completely unavailable (indoor or covered areas and underpasses).
For this reason, a lot of effort is directed towards the development of map-based localization algorithms \cite{fox_kld-sampling_2002,hess_real-time_2016}, that compare live sensor data (typically LiDARs or cameras) to an a-priori known map of the environment to estimate the vehicle location.
While these provide invaluable data in some settings, they are not a silver bullet and previous works have shown how they may struggle in some scenarios like vast open areas and repetitive environments \cite{mozzarelli_comparative_2023}.
To mitigate these issues sensor fusion has been explored since the early days of robotics. 
Although some modern approaches integrate more sensors, like GNSS receivers, in a graph Simultaneous Localization and Mapping (SLAM) framework \cite{shan_lio-sam_2020}, these tend to be computationally complex and rarely consider the peculiarities of the vehicle on which the algorithm is deployed, resulting in high sensitivity to sensor calibrations and completely unrealistic motion estimates during failures (sudden jumps, lateral translations for holonomic vehicles or vertical movement for robots working in a planar environment, for example).
Indeed, most of these SLAM systems aim to be off-the-shelf algorithms compatible with a vast variety of vehicles and setups, assuming unconstrained rigid body dynamics.
Adapting them to exploit the motion constraints imposed by a specific vehicle can be a resource-intensive hurdle, even though the benefits could be noteworthy.
More classical approaches employ Extended Kalman Filters (EKFs), as it is a lightweight and versatile sensor fusion algorithm, with classical examples fusing proprioceptive sensors (like IMUs and wheel encoders) with GNSS measurements \cite{boberg_integrating_2002,rezaei_kalman_2007}.
This enables the creation of high-frequency and drift-free positional data as long as an unobstructed view of the sky is available which, as previously stated, is a strong assumption.
While approaches fusing a vast variety of sensors have been developed over the years \cite{sabatini_sensor_2014,pizzocaro_magnetometer_2021}, GNSS receivers have long remained the only source of absolute localization data, with few application and environment-specific exceptions \cite{wischnewski_vehicle_2019}.

This paper proposes a modular, two-layered localization architecture that can incorporate raw sensor measurements, as well as the output of off-the-shelf localization algorithms.
This enables the exploitation of complex, vehicle-agnostic algorithms in a localization system that is aware of the vehicle's peculiarities, with relevant benefits in the reliability of odometrical estimates.
Indeed, while in \cite{wischnewski_vehicle_2019} little thought is given to the availability of the position measurements, in an urban area position sources might unexpectedly become unavailable: GNSS measurements may be lost for the already mentioned multi-path and sky coverage issues, while map-based methods might fail due to changes in the environment or challenging locations.
Such losses would force to robot to halt, even if continuing along the trajectory would allow it to re-localize (think of an underpass that blocks GNSS reception: reaching the end would allow the reception of new measures).
While fusing two position sources definitely increases the robustness of the localization solution, loss of both is not uncommon.
In such conditions, having short or medium-term resilience to complete measurement loss can significantly increase the robot's autonomous capabilities and reduce the need for human intervention.
Since proprioceptive sensors are highly reliable, continuing to move trusting an odometrical estimate is an option but, without the necessary precautions, it is often too risky: the drift rates of odometry can be unpredictable and highly affected by model parameters like wheel radii.
Our approach improves the odometrical estimation by estimating model uncertainties online, resulting in relevant improvements in positional errors.

To recap, the main contribution of this paper is the development of a localization architecture which is:
\begin{itemize}
  \item modular and effortlessly expandable to incorporate new localization sources, both physical sensors and off-the-shelf algorithms;
  \item aware of vehicle-specific motion models,
  \item which enables the improvement of odometrical data by estimating model uncertainties;
\end{itemize}
The architecture is also capable of handling asynchronous, multi-rate and unavailable measurements by adapting the estimated state based on time-varying observability conditions and can be easily ported to different vehicles and setups.

The remainder of this paper is structured as follows.
Section \ref{sec:architecture} discusses the core principles of the proposed localization architecture.
Next, Section \ref{sec:setup} will present the robotic platform used to develop and validate the proposed algorithm.
The implementation of the architecture for the specific experimental platform is detailed in \ref{sec:yape_architecture} and the experimental results are reported in Section \ref{sec:validation}.

\section{LOCALIZATION ARCHITECTURE}
\label{sec:architecture}
\begin{figure}[!t]
  \centering
  \includegraphics[width=0.49\textwidth]{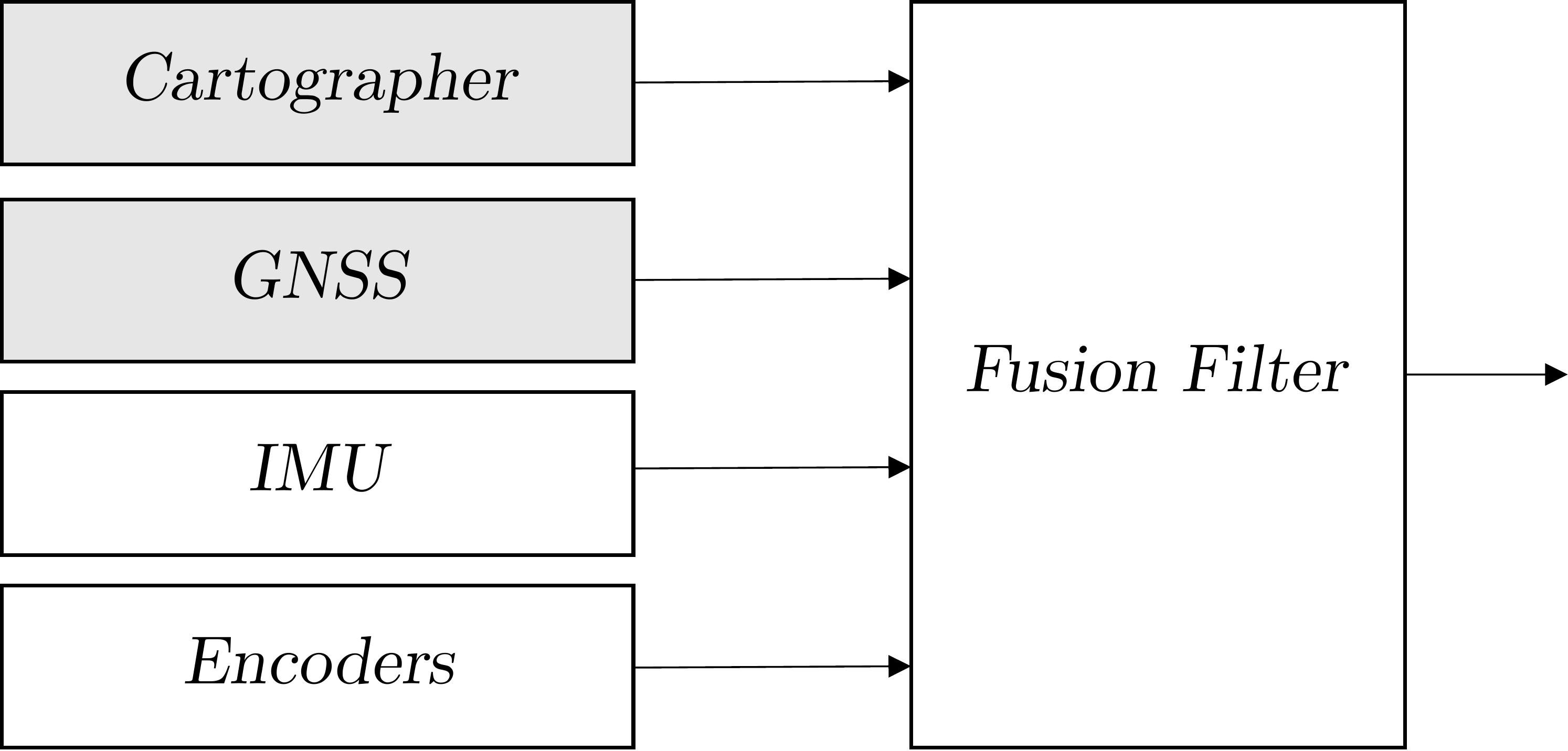}
  \caption{Block scheme of the proposed localization architecture.}
  \label{fig:architecture}
\end{figure}

The proposed localization architecture comprises two layers, exemplified in Figure \ref{fig:architecture}.
The first layer, on the left, is constituted by the localization sources, while the second layer performs the sensor fusion operations.
Localization sources could be physical sensors, like GNSS receivers, encoders, IMUs and magnetometers or other algorithms that provide useful localization data, like map-based localization algorithms or LiDAR odometries.
In general, three categories can be identified:
\begin{itemize}
  \item Absolute pose (or position) sources provide the robot's position and, optionally, orientation with respect to a fixed reference frame.
  \item Displacement sources include odometries based on exteroceptive sensors, like LiDAR or visual odometries.
  \item Proprioceptive sensors measure the internal state of the vehicle, like IMUs and encoders.
\end{itemize}
Two remarks are in order regarding the interaction between the two layers.
Firstly, it is reasonable to expect that different sources will have different sampling frequencies and the second layer must be capable of handling such mismatch.
Secondly, when multiple absolute position sources are considered, each might provide data in a different reference frame.
For the fusion to work correctly, a calibration process to estimate the roto-translation between such references is necessary.
We solve this issue by collecting raw data from the different sources in a first mapping experiment.
The data can then be post-processed to extract the trajectory segments in which both sources provide reliable measurements and the corresponding trajectories get aligned by minimizing the point-to-point mean squared error.
The minimization can be performed in closed form by means of the method by Horn \cite{horn_closed-form_1988}.

The second layer of the proposed localization architecture consists of a filter based on the Extended Kalman Filter framework.
Under this extremely popular framework, a model is employed to predict the evolution of the state between iterations and the prediction is subsequently corrected by integrating measures.
The applied correction is proportional to the error between measured values and measurement predictions, obtained through an observation model from the predicted state.
In the proposed approach, and contrary to \cite{wischnewski_vehicle_2019}, we advocate for vehicle-specific models.
This enables the use of formulations that are better tailored: the more accurate the state transition model is the more robust the localization will be to failures in the absolute localization error sources.

As noted above, the filter must be capable of handling a varying number of measurements, which is not a given in the EKF paradigm.
A common method to bypass this issue to handle multi-rate measurements is to design the filter to run at the frequency of the slowest sensor.
This solution is clearly suboptimal and might not meet the output frequency requirements.
Furthermore, some of the inputs might become unavailable for prolonged periods: GNSS data might not be reliable if the robot is traveling indoors, while map-based localization methods might lose their lock with the localization map if the environment presents substantial changes.
For these reasons, we approach this issue differently: selecting the filter execution frequency based on the fastest sensor.
At each iteration, we perform the prediction step and the correction step, but the latter considers only the available measurements.
This is achieved by dynamically adjusting the dimensions of the involved matrices.
Notice that, depending on the formulation of the model and the redundancy of the sensors, the observability properties of the system might change.
A method to handle this situation will be presented in Section \ref{sec:yape_architecture/observability}.

\section{EXPERIMENTAL SETUP}
\label{sec:setup}
\begin{figure}[!t]
  \centering
  \includegraphics[width=0.3\textwidth]{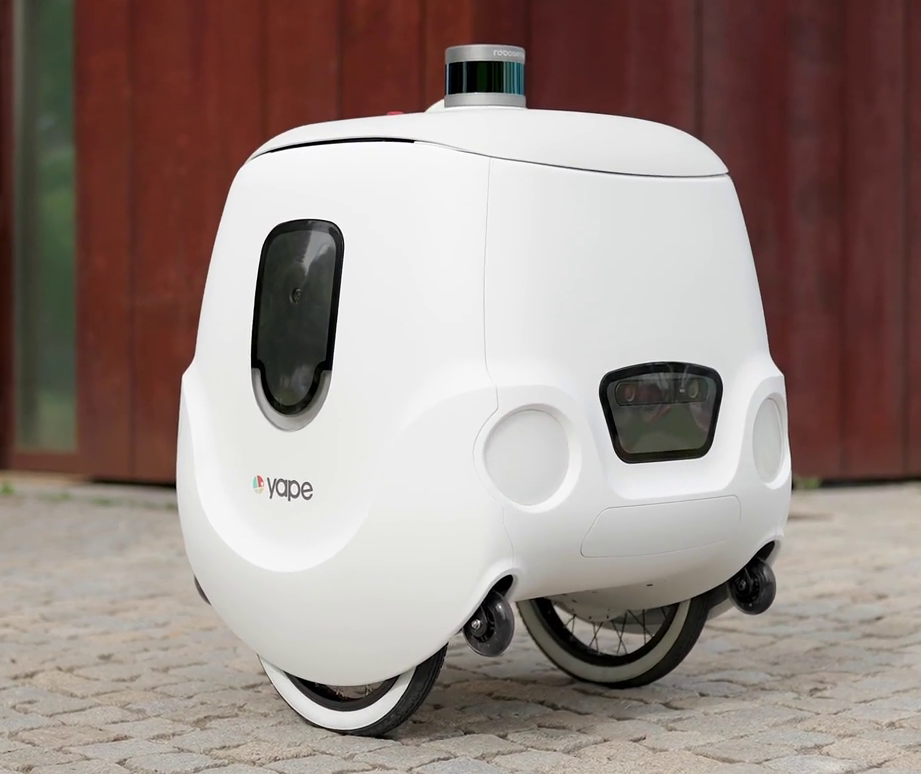}
  \caption{Yape, the deployment target of the developed localization algorithm.}
  \label{fig:yape}
\end{figure}
The localization architecture presented in this paper was developed and tested on Yape, an autonomous ground drone designed for last-mile delivery (\cite{sabatini_improving_2018,sabatini_vision-based_2018,parravicini_robust_2019,parravicini_extended_2020,corno_measuring_2020}).
Yape (Figure \ref{fig:yape}) is a Two-Wheeled Inverted Pendulum (TWIP) robot, which independently controls the two wheels to achieve motion while keeping the chassis balanced.
The robot is equipped with a sensor package comprising an Ardusimple RTK GNSS receiver, a Robosense R16 3D LiDAR, a chassis-mounted 6 degrees of freedom Inertial Measurement Unit (IMU) and two wheel encoders.
The IMU measures linear accelerations and rotational velocities along the three axis, while wheel encoders provide wheel positions and rotational speeds.
The presented algorithm was implemented running online on the onboard processing unit, an Nvidia Xavier AGX development board, by employing the ROS framework \cite{quigley_ros_2009}.

\section{ARCHITECTURE SPECIALIZATION FOR YAPE}
\label{sec:yape_architecture}
This Section details how the architecture principles outlined in Section \ref{sec:architecture} translate in a practical implementation on the Yape AMR.

\subsection{Localization sources}
\label{sec:yape_architecture/sources}
In this work, we fuse data from two types of sources: absolute pose sources and proprioceptive sources, leaving the integration of displacement sources to future develoments.
To achieve absolute localization, the GNSS receiver provides position measures at $10~ Hz$.
GNSS data achieves centimeter-level accuracy when a clear view of the sky and an internet connection to receive RTK corrections are available.
Furthermore, a LiDAR, map-based localization algorithm was deployed to provide absolute pose data indoors and under poor GNSS coverage.
Cartographer \cite{hess_real-time_2016} was selected to cover this role after a detailed analysis of its performance, presented in \cite{mozzarelli_comparative_2023}.
The Cartographer algorithm is employed in pure localization mode: the localization map is considered available after a mapping experiment.
In said experiment, the robot is teleoperated on the area that will be autonomously traversed while acquiring raw sensor data.
The dataset is then fed to the Cartographer algorithm in SLAM mode, which builds and optimizes the map.
Note that GNSS and Cartographer data are referred to different reference frames: to estimate the roto-translation between the two, the mapping phase is exploited, applying the alignment method described in Section \ref{sec:architecture} to the optimized SLAM trajectory and the GNSS one.
Cartographer pose data is provided at a frequency of $20~Hz$.
Regarding proprioceptive sources, we fuse the wheel speeds measured by the encoders, as well as the yaw rate measurement from the IMU.
Both sensors sample data at $100~Hz$.

\subsection{State transition model}
\label{sec:yape_architecture/state_model}
\begin{figure*}[h!]
  \centering
  \subfloat[]{\includegraphics[width=0.48\textwidth]{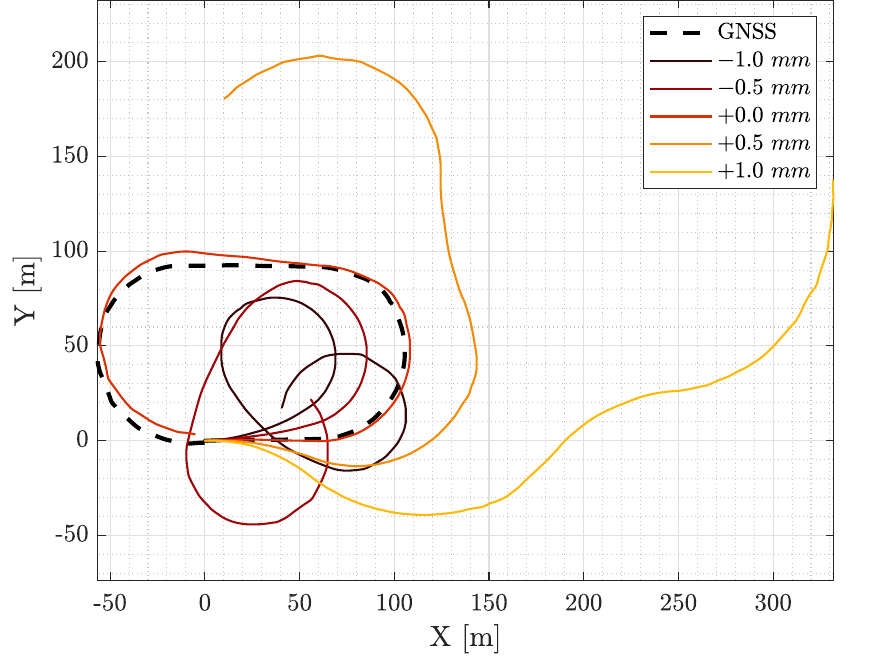}%
  \label{fig:yape_arch/odometry_radius_sensitivity}}
  \hfill%
  \subfloat[]{\includegraphics[width=0.48\textwidth]{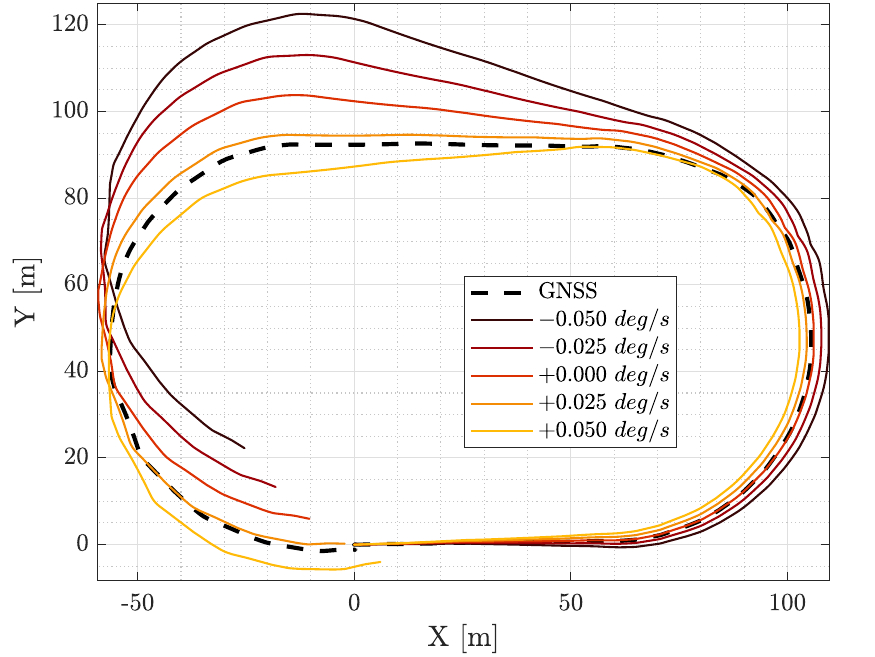}%
  \label{fig:yape_arch/odometry_imu_bias_sensitivity}}
  \caption{Sensitivity analysis of model uncertainties on the estimated (odometrical) trajectory: left wheel radius \protect\subref{fig:yape_arch/odometry_radius_sensitivity} and IMU yaw rate bias \protect\subref{fig:yape_arch/odometry_imu_bias_sensitivity}.}
  \label{fig:yape_arch/odometry_uncert_sensitivity}
\end{figure*}

To show the potential of implementing vehicle-specific models we developed an uncertain kinematic differential model.
The estimated state $x$ is
\begin{equation}
  x_k = 
  \begin{bmatrix}
  X_k & Y_k & \psi_k & \omega^R_k & \omega^L_k & R^R_k & R^L_k & b_k
  \end{bmatrix}^T
  \label{eq:filter/state}
\end{equation}
where $X$, $Y$ and $\psi$ represent the pose of the robot, $\omega^R$ and $\omega^L$ the wheels rotational speeds while $R^R$, $R^L$ and $b$ the wheel radii and IMU yaw rate bias.
Subscript $k$ indicates that the variables are referred to time instant $k$.
The last three states represent uncertain model parameters, which strongly affect the odometrical performance of the localization solution.
A sensitivity analysis of this phenomenon is reported in Figure \ref{fig:yape_arch/odometry_uncert_sensitivity}.
Notice how even small inaccuracies can have huge effects on the estimated trajectory.

We make the assumption of low wheel speeds, radii and IMU bias dynamics, which allows us to write the state transition function as
\begin{equation}
  x_{k+1} = f(x_k,u_k) = 
  \begin{bmatrix}
    X_{k} + \frac{\omega^R_k R^R_k + \omega^L_k R^L_k}{2} \cos{\psi_k} T_s \\
    Y_{k} + \frac{\omega^R_k R^R_k + \omega^L_k R^L_k}{2} \sin{\psi_k} T_s \\
    \psi_k + \frac{\omega^R_k R^R_k - \omega^L_k R^L_k}{T} T_s \\
    \omega^R_k \\
    \omega^L_k \\
    R^R_k \\
    R^L_k \\
    b_k
  \end{bmatrix}
  \label{eq:filter/state_transition_function}
\end{equation}
where $T_s$ is the sampling time and $T$ the vehicle's track width.
Note that in our formulation $u_k=\emptyset$: we include all sensor data as outputs. 

\subsection{Measurement model}
\label{sec:yape_architecture/measures}
Since Cartographer and encoder measurements map directly to state variables the predicted measures $\hat y$ will be
\begin{equation}
  \hat y_k^{enc} = \begin{bmatrix}
      \omega_k^R \\ \omega_k^L
  \end{bmatrix}
\end{equation}
and
\begin{equation}
  \hat y_k^{carto} = \begin{bmatrix}
      X_k \\ Y_k \\ \psi_k
  \end{bmatrix}.
  \label{eq:filter/measures/carto}
\end{equation}
The measurement model of the GNSS receiver is similar to \eqref{eq:filter/measures/carto} but lacks a heading measurement and needs to take into account the displacement of the antenna mounting position with respect to the vehicle's center of mass, expressed here in polar coordinates $(d,\alpha)$:
\begin{equation}
  \hat y_k^{gnss} = \begin{bmatrix}
    X_k + d \cos (\psi_k+\alpha) \\ Y_k + d \sin (\psi_k+\alpha)
  \end{bmatrix}.
\end{equation}
Finally, the IMU yaw rate measurement prediction needs to be computed from both wheel speeds, both radii as well as the bias:
\begin{equation}
  \hat y_k^{imu} = \frac{\omega^R_k R^R_k - \omega^L_k R^L_k}{T} + b_k .
  \label{eq:measurement_model_imu}
\end{equation}

\subsection{Dealing with multi-rate and unavailable measures}
\label{sec:yape_architecture/observability}
As introduced above, we aim to achieve a high-frequency output, independently from the acquisition frequency of the slowest sensor.
In our case, both the encoders and the IMU provide data at $100~Hz$: we select the execution frequency $1/T_s=60~Hz$ to maximize the probability of receiving new samples from the proprioceptive sensors before each iteration, even under severe congestion of the ROS network.
In practice, this means that the predicted measurements vector will have a minimum of 3 elements and a maximum of 8, with GNSS measures being integrated roughly every 6 iterations and Cartographer ones every 3.

An important side-effect of this design choice is that the observability properties of the system change between one iteration and the following one.
Indeed, in all the iterations where no position measurements are available, the linearized measurement matrix would result in being
\begin{equation}
  C = \begin{bmatrix} 0 & 0 & 0 & \frac{R_{R}}{T} & -\frac{R_{L}}{T} & \frac{\omega _{R}}{T} & \frac{\omega _{L}}{T} & 1\\ 0 & 0 & 0 & 1 & 0 & 0 & 0 & 0\\ 0 & 0 & 0 & 0 & 1 & 0 & 0 & 0 \end{bmatrix} .
\end{equation}
The $3 \times 3$ null block will cause the first three columns of the observability matrix $\mathcal O = \begin{bmatrix} C & CA & \dots & CA^7 \end{bmatrix}^T$ to be zero-valued as well.
As a matter of fact, the number of linearly independent rows would be $\rank \mathcal{O}=3$: the observable state is reduced to the two wheel speeds and one between the radii and the yaw rate bias.

Some works in the literature solve this issue by predicting the measurement as usual and inserting it as a virtual measurement \cite{katriniok_adaptive_2016} or by dividing the estimation problem into multiple sub-filters \cite{bristeau_design_2010}.
The former approach, however, induces an artificial convergence of the unobservable states covariances. The latter, while being theoretically sounder, cannot consider cross-correlations between the states of different filters and its implementation is inevitably complex.
We opt for a different approach, which applies the prediction steps of the EKF to the full state, while the correction step is applied only to the observable states.
Although one of the model uncertainty states would theoretically be observable even without position measures, we choose to correct just the wheel speeds.
Indeed if, as an example, we opted to estimate the IMU bias as well, the  measurement model in \eqref{eq:measurement_model_imu} would become
\begin{equation}
  \hat y_k^{imu} = \frac{\omega^R_k \bar R^R - \omega^L_k \bar R^L}{T} + b_k
\end{equation}
where $\bar R^R$ and $\bar R^L$ are the latest available estimates of the wheel radii.
However, it is not possible to assert that such estimates have reached convergence, and employing erroneous values would result in an incorrect estimate of the bias $b$ as well.

In practice, this approach results in the model uncertainty states remaining constant, while the position states get updated in an odometrical fashion.
Notice that the covariance of the unobservable states grows during these \quotes{open-loop} periods, since the state covariance prediction step is applied to the full state.
\section{VALIDATION RESULTS}
\label{sec:validation}
\begin{figure}[t!]
  \centering
  \vspace{-2.2em}
  \includegraphics[width=0.48\textwidth]{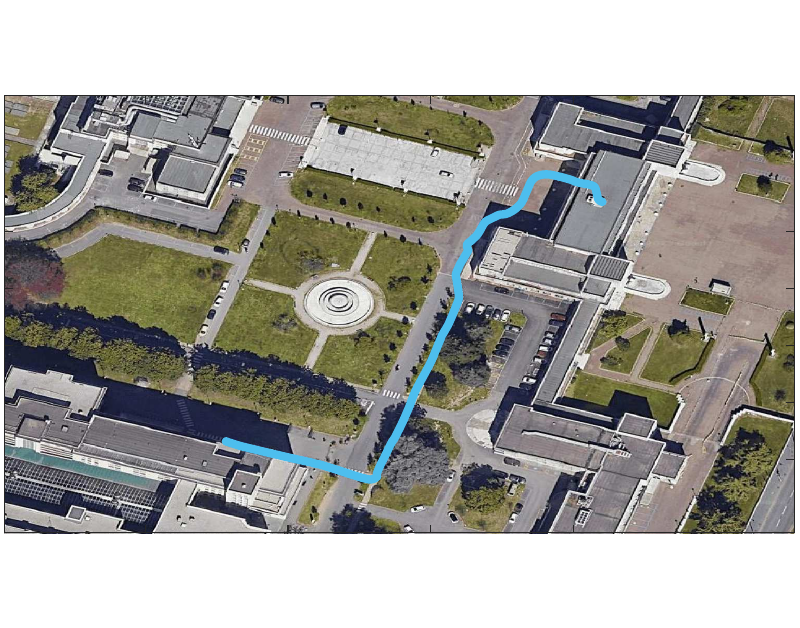}%
  \vspace{-3.7em}
  \caption{Aerial view of the trajectory traveled by Yape in the courtyard of the Niguarda Hospital in Milan.}
  \label{fig:validation/20230512_niguarda_trajectory}
\end{figure}

The proposed algorithm has been extensively tested in field trials in which the robot autonomously navigated public sidewalks.
The experiment reported in this paper was conducted in the spacious courtyard of the Niguarda Hospital (\textit{ASST Grande Ospedale Metropolitano Niguarda}) in Milan during test runs of the Budd-e project \cite{rebecchi_shaping_2023}.
The traversed courtyard is a vast area with heavy pedestrian and vehicular traffic, presenting numerous buildings and green areas.
Figure \ref{fig:validation/20230512_niguarda_trajectory} shows the aerial view of the robot trajectory.

The presented analysis focuses on the robustness of the localization solution to the loss of absolute position measurements and the effect of model uncertainty estimation.
The first set of experiments proves the robustness of the algorithm to the loss of one of the two absolute pose measurements, either the GNSS or Cartographer.
The second analysis presents the effect of model uncertainty estimation with position measurement available.
The last set of experiments combines the previous two, by assessing the performance when all position measures are lost and evaluating the impact of estimating model uncertainties under such circumstances.

\subsection{Loss of one absolute pose source}
\begin{figure}[h!]
  \centering
  \includegraphics[width=0.48\textwidth]{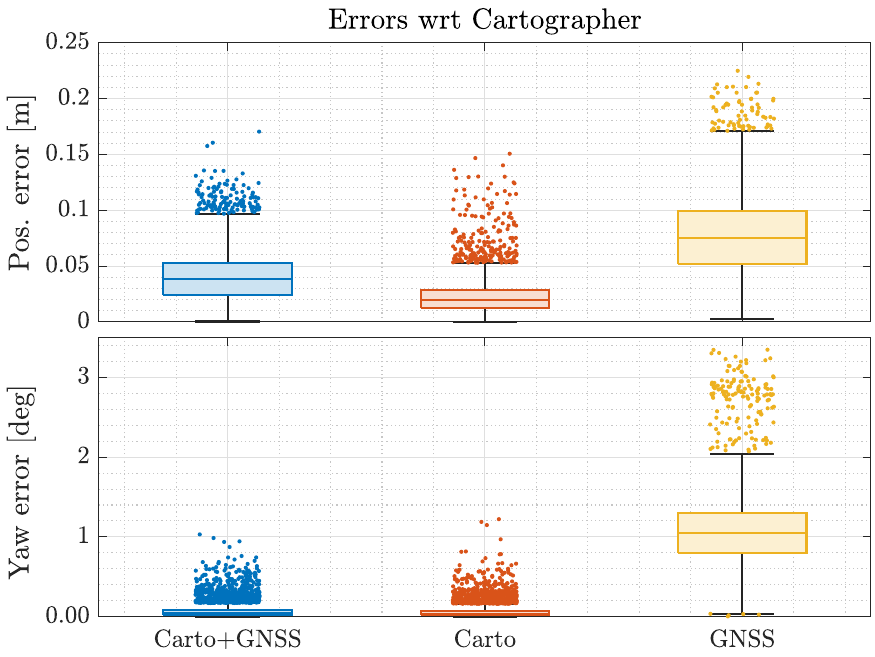}%
  \caption{Position (top) and yaw (bottom) error distribution of filter estimate with respect to Cartographer. The X axis refers to the fused position inputs (GNSS, Cartographer or both).}
  \label{fig:validation/20230512_niguarda_errors_wrt_carto}
\end{figure}

\begin{figure}[h!]
  \centering
  \includegraphics[width=0.48\textwidth]{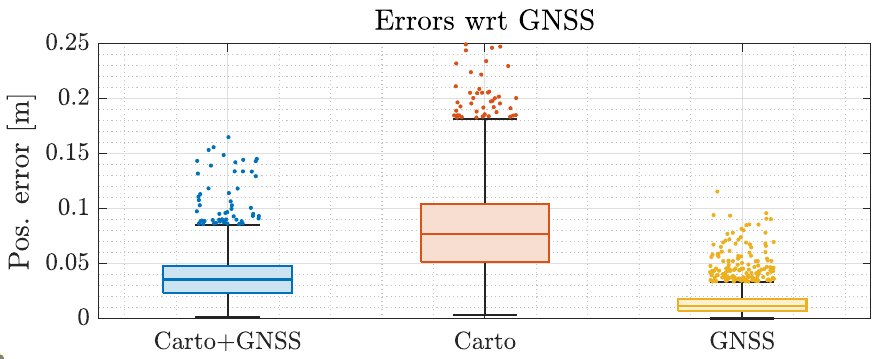}%
  \caption{Position error distribution of filter estimate with respect to GNSS. The X axis refers to the fused position inputs (GNSS, Cartographer or both).}
  \label{fig:validation/20230512_niguarda_position_errors_wrt_gnss}
\end{figure}

In the presented experiment, both GNSS and Cartographer provide high-quality measurements during the entirety of the test, exception made for the final seconds in which GNSS quality degrades as the robot enters a indoor area (top right in Figure \ref{fig:validation/20230512_niguarda_trajectory}).
This allowed us to simulate the loss of one sensor by disabling the fusion in the filter while using it for ground truth purposes.
Figures \ref{fig:validation/20230512_niguarda_errors_wrt_carto} and  \ref{fig:validation/20230512_niguarda_position_errors_wrt_gnss}  present the results of these experiments: in the former, the error between the output of the filter and Cartographer measures remains close to the raw sensor data, even when fed GNSS data only (yellow plots).
Note that in spite of not having direct yaw measurements, the error maintains a median of $1 \deg$.
Similar conclusions can be drawn by inspecting the latter, which reports errors with respect to the GNSS position.

\subsection{Model uncertainties estimation}
\begin{figure}[h!]
  \centering
  \includegraphics[width=0.48\textwidth]{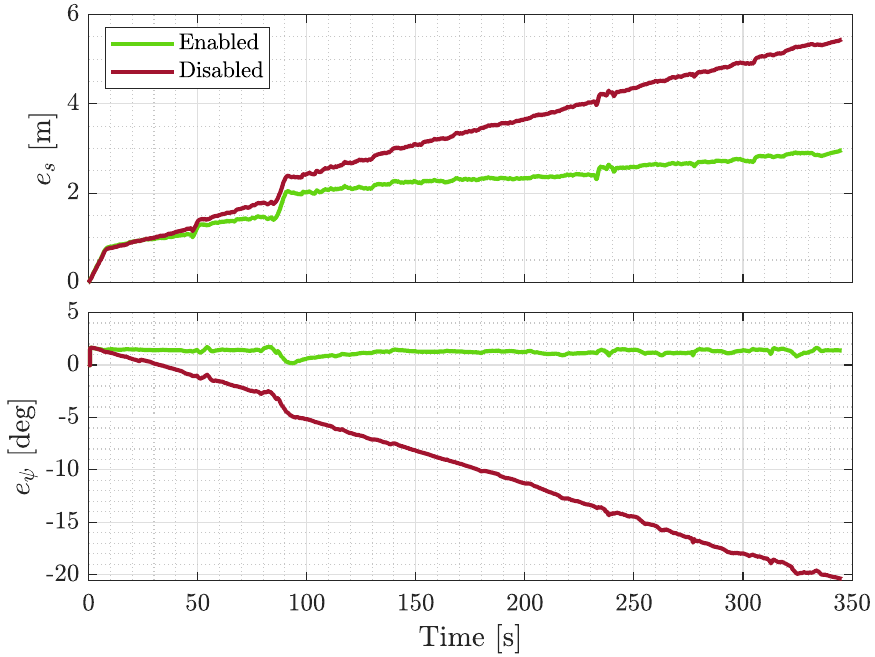}%
  \vspace{-1.5em}
  \caption{Effect of model uncertainties estimation on the curvilinear abscissa (top) and yaw (bottom) error between the filter output pose and integrated output velocities. The experiment was repeated with (green lines) and without (red line) model uncertainties estimation.}
  \label{fig:validation/20230512_niguarda_uncert_estimation_cumulated_error}
\end{figure}

The next set of experiments shows the effect of the model uncertainty estimation on the velocity states.
The goal of this analysis is to show how estimating the wheel radii and IMU yaw rate bias is necessary to achieve coherency between the position and velocity states of the filter.
Two experiments were performed, with and without model uncertainty estimation, and for each one the position outputs were used to compute the cumulative traveled distance $s$.
Similarly, the output velocities were integrated to obtain both the traveled distance and cumulated yaw angle.
The top plot in Figure \ref{fig:validation/20230512_niguarda_uncert_estimation_cumulated_error} displays the error between the cumulated distance  computed from the poses and the one computed from the velocities, while the bottom one shows the error between the yaw angle estimated from the EKF and the one obtained by integrating the estimated yaw rate.
With the model uncertainty estimation disabled (dark red lines) both errors diverge: this means that the velocities are not getting estimated correctly, and their integration drifts.
With the estimation enabled (light green lines), after a convergence phase, such drift is significantly reduced from the longitudinal distance and even eliminated entirely from the yaw angle.

\subsection{Loss of all absolute pose sources}
\begin{figure}[h!]
  \centering
  \subfloat[]{\includegraphics[width=0.48\textwidth]{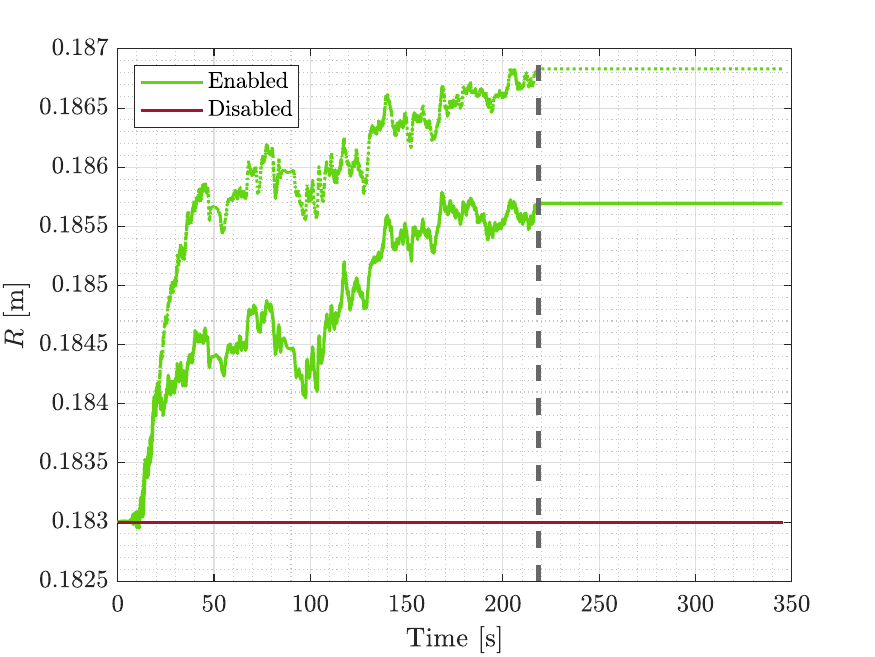}%
  \label{fig:validation/20230512_niguarda_stop_pose_meas_radii}}
  \hfil%
  \subfloat[]{\includegraphics[width=0.48\textwidth]{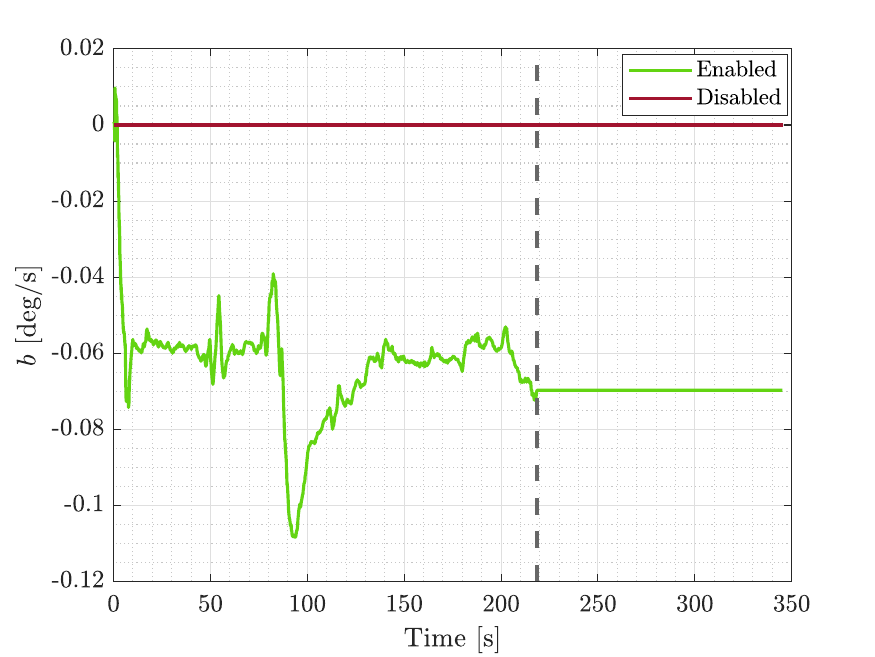}%
  \label{fig:validation/20230512_niguarda_stop_pose_meas_imu_bias}}
  \caption{Result of model uncertainties estimation when both position measures are temporarily lost: wheel radii \protect\subref{fig:validation/20230512_niguarda_stop_pose_meas_radii}  (continuous and dotted lines represent the left and right wheel, respectively) and IMU yaw rate bias \protect\subref{fig:validation/20230512_niguarda_stop_pose_meas_imu_bias}. The experiment was repeated with (green lines) and without (red line) model uncertainties estimation.}
  \label{fig:validation/20230512_niguarda_stop_pose_meas_uncert_states}
\end{figure}

\begin{figure*}[h!]
  \centering
  \subfloat[]{\includegraphics[width=0.48\textwidth]{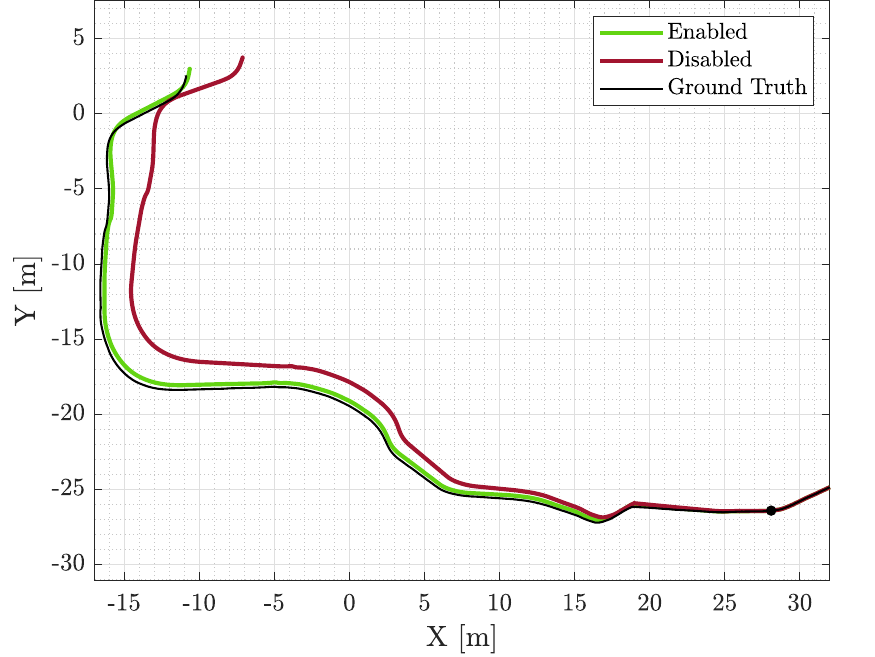}%
  \label{fig:validation/20230512_niguarda_stop_pose_meas_traj}}
  \hfill%
  \subfloat[]{\includegraphics[width=0.48\textwidth]{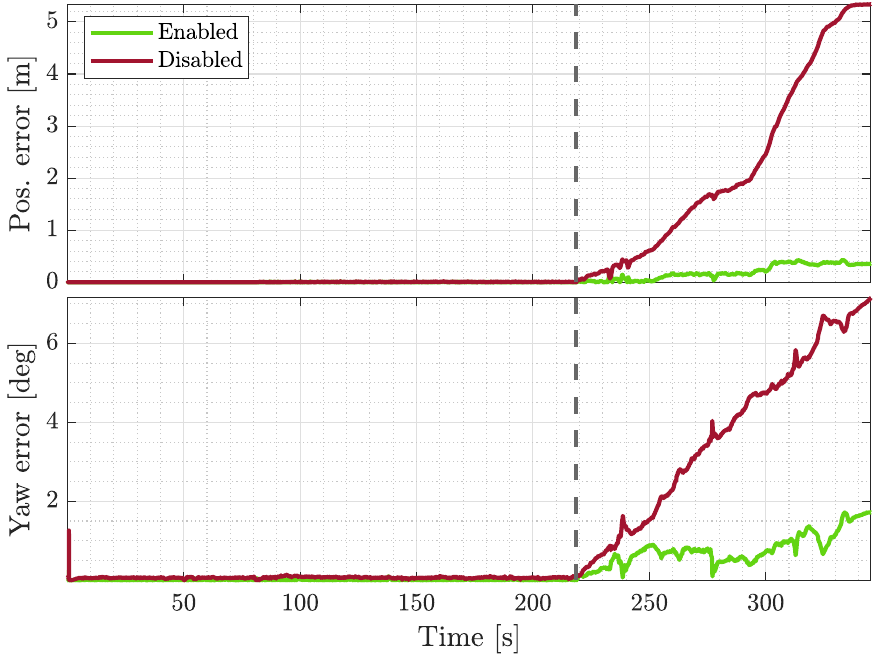}%
  \label{fig:validation/20230512_niguarda_stop_pose_meas_errors}}
  \caption{Effect of model uncertainties estimation on the estimated (odometrical) trajectory \protect\subref{fig:validation/20230512_niguarda_stop_pose_meas_traj}when position measures are lost: position (\protect\subref{fig:validation/20230512_niguarda_stop_pose_meas_errors}, top)  and yaw (\protect\subref{fig:validation/20230512_niguarda_stop_pose_meas_errors}, bottom) errors. The black dot in \protect\subref{fig:validation/20230512_niguarda_stop_pose_meas_traj} and dashed lines in \protect\subref{fig:validation/20230512_niguarda_stop_pose_meas_errors} indicate the last position measurement received. The experiment was repeated with (green lines) and without (red line) model uncertainties estimation.}
\end{figure*}

The final validation experiments aim to assess the effect of losing both position measures, a severe but not uncommon circumstance in real-world deployments of autonomous vehicles.
To simulate this scenario, both the GNSS and Cartographer fusion are disabled from $t=220~s$, leaving the algorithm running with proprioceptive inputs only for $130 ~s$.
The trajectory estimated by the filter fed with both inputs was employed as ground truth.
Figure \ref{fig:validation/20230512_niguarda_stop_pose_meas_uncert_states} shows the values of the estimated radii and bias.
Notice how the loss of position measurements (vertical dashed line) stops their estimation since we assume their evolution to be static.
The effect of the estimation is clearly visible in Figure \ref{fig:validation/20230512_niguarda_stop_pose_meas_traj}, which depicts the trajectories after the measurement loss occurs (black dot).
With the model uncertainties estimated until such moment, the resulting model is much more accurate and enables to maintain significantly lower errors and minimize odometrical drift, as visible in Figure \ref{fig:validation/20230512_niguarda_stop_pose_meas_errors}, which reports the time series of the filter output with respect to Cartographer's pose.
Notice how, after more than two minutes of navigation and $\approx 70~m$ traveled the cumulated drift reduces from $5.3~m$ to $0.35~m$ in terms of position error and from $7~\deg$ to less than $2~\deg$ in terms of yaw error.

\section{CONCLUSIONS}
\label{sec:conclusions}
This paper presented a localization architecture for autonomous mobile robots.
Structured on two layers, the first one encompasses localization sources including both physical sensors (IMU, encoders and GNSS) and fully-featured off-the-shelf localization algorithms (Cartographer).
The second layer is a fusion filter based on the Extended Kalman Filter framework and tailored to the vehicle motion model, allowing the estimation of model uncertainties and the subsequent improvement of odometry estimates.
This represents a relevant improvement with respect of vehicle-agnostic localization algorithms, which despite their undoubtable usefulness, can fail in unrealistic ways.
The filter is capable of handling multi-rate and unavailable measurements by adapting the estimated state online based on to the time-varying observability properties.
Robustness to complete loss of position measurements is demonstrated by disabling Cartographer and GNSS corrections for more than two minutes, resulting in a cumulated error of $0.35~m$, with respect to the $5.3~m$ achieved without uncertainties estimation.

\bibliography{IEEEabrv,bibliography}
\bibliographystyle{IEEEtran}

\end{document}

%% file: custom_commands.tex
\usepackage{siunitx}					
\usepackage{eurosym}					
\usepackage{graphicx}
\usepackage[caption=false,font=normalsize,labelfont=sf,textfont=sf]{subfig}
\usepackage{multirow}
\usepackage{amsmath}
\usepackage{amsfonts}
\usepackage{bm}

\newcommand{\quotes}[1]{``#1''}
\newcommand*\xbar[1]{%
  \hbox{%
    \vbox{%
      \hrule height 0.5pt 
      \kern0.5ex
      \hbox{%
        \kern-0.1em
        \ensuremath{#1}%
        \kern-0.1em
      }%
    }%
  }%
}

\makeatletter
\DeclareRobustCommand{\atan}{%
  \operatorname{atan}%
  \@ifnextchar2{_}{}%
}
\makeatother

\DeclareMathOperator*{\rank}{rank}

\DeclareSIUnit{\pers}{pers}
\DeclareSIUnit{\EUR}{\text{\euro}}
